\documentclass[10pt,twocolumn,letterpaper]{article}

\usepackage[T1]{fontenc}

\usepackage{cvpr}
\usepackage{times}
\usepackage{epsfig}
\usepackage{graphicx}
\usepackage{amsmath}
\usepackage{amssymb}
\usepackage[ruled,noend]{algorithm2e}
\usepackage{amsmath,amssymb,amsfonts,amsbsy,amsfonts,latexsym}
\usepackage{multirow}
\usepackage{makecell}
\usepackage{subcaption}
\usepackage[labelfont=bf,textfont=it,belowskip=0pt,aboveskip=5pt,tableposition=top]{caption}
\usepackage[dvipsnames]{xcolor}
\usepackage{colortbl}
\usepackage{booktabs}
\newcounter{runidnum}

\definecolor{colorA}{RGB}{189,201,225}
\definecolor{colorB}{RGB}{103,169,207}
\definecolor{colorC}{RGB}{ 28,144,153}
\definecolor{colorD}{RGB}{  1,108, 89}

\newcolumntype{R}{>{\columncolor{gray!40}}r}
\newcolumntype{L}{>{\columncolor{gray!40}}l}
\newcolumntype{C}{>{\columncolor{gray!40}}c}

\newcommand\fref{Fig.~\ref}
\newcommand\tref{Table~\ref}
\usepackage[breaklinks=true,bookmarks=false]{hyperref}

\usepackage{xparse}
\NewDocumentCommand{\var}{O{s} m O{}}{%
  \ensuremath{#1_{#2}^{#3}}
}

\newcommand{\R}{\mathcal{R}}

\definecolor{light-gray}{gray}{0.80}

\renewcommand\paragraph{\subsubsection*}

\newcommand\ga{\rowcolor{gray!0}}

\newcommand\gc{\rowcolor{gray!30}}

\newcommand\cmt[1]{\tcp*[r]{\scriptsize \color{gray!80!black}#1}}

\newcommand{\pexp}{\text{\sc{e}{\scriptsize +}}} 

\graphicspath{{../}}
\def\cvprPaperID{****} 

\cvprfinalcopy
\begin{document}

\title{SqueezeNext: Hardware-Aware Neural Network Design}


\author{Amir Gholami, Kiseok Kwon, Bichen Wu, Zizheng Tai,\\ Xiangyu Yue, Peter Jin, Sicheng Zhao, Kurt Keutzer
 \\
EECS, UC Berkeley\\
{\tt\small \{amirgh,kiseok.kwon,bichen,zizheng,xyyue,phj,schzhao,keutzer\}@berkeley.edu}
}

\maketitle

\begin{abstract}
  One of the main barriers for deploying neural networks on embedded systems has been large memory and power consumption
  of existing neural networks.
  In this work, we introduce SqueezeNext, a new family of neural network architectures whose design was guided by considering
  previous architectures such as SqueezeNet, as well as by simulation results on a neural network accelerator.
  This new network is able to match AlexNet's accuracy on the ImageNet benchmark with $112\times$ fewer parameters,
  and one of its deeper variants is able to achieve VGG-19 accuracy with only 4.4 Million parameters, ($31\times$ smaller than VGG-19).
  SqueezeNext also achieves better top-5 classification accuracy with $1.3\times$ fewer parameters as compared to MobileNet, but avoids using depthwise-separable convolutions that are inefficient on some mobile processor platforms. 
  This wide range of accuracy gives the user the ability to make speed-accuracy tradeoffs, depending on the available resources on the target hardware.
  Using hardware simulation results for power and inference speed on an embedded system has guided us to design variations of the baseline model that are $2.59\times$/$8.26\times$
    faster and $2.25\times$/$7.5\times$ more energy efficient as compared to SqueezeNet/AlexNet without any accuracy degradation.
\end{abstract}

\section{Introduction}
\label{s:intro}

\begin{figure*}[!htbp]
\centering
\includegraphics[width=0.7\textwidth]
{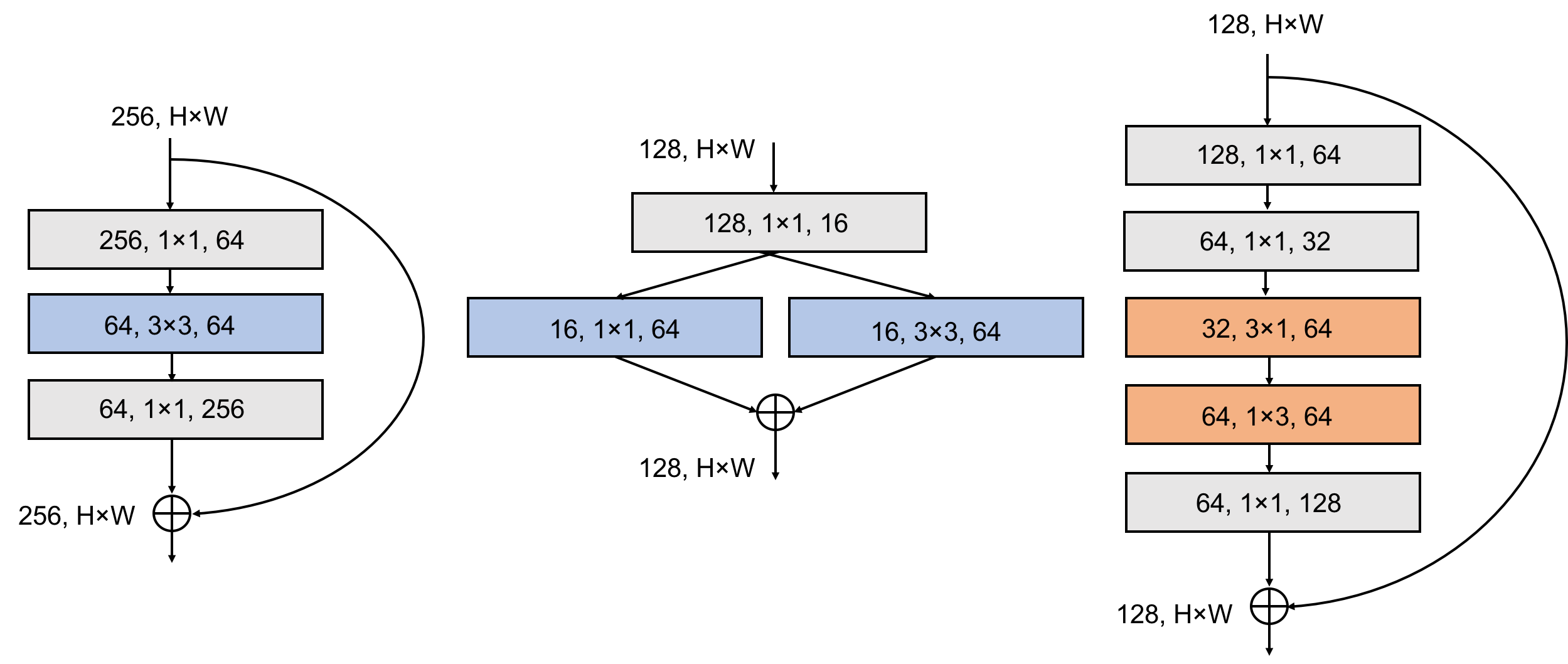}
\caption{
  Illustration of a ResNet block on the left, a SqueezeNet block in the middle, and a SqueezeNext (SqNxt) block on the right.
  SqueezeNext uses a two-stage bottleneck module to reduce the number of input channels to
  the $3\times3$ convolution. The latter is further decomposed into separable convolutions to further reduce the number of parameters (orange parts), followed by a $1\times1$ expansion module.
  }
\label{f:separable}
\end{figure*}

Deep Neural Networks have transformed a wide range of applications in computer vision.  This has been made
possible in part by novel neural net architectures, the availability of more training data, and faster hardware for both training and inference. 
The transition to Deep Neural Network based solutions started with
AlexNet~\cite{krizhevsky2012imagenet}, which won the ImageNet challenge by a large margin.
The ImageNet classification challenge started in 2010 with the first winning
method achieving an error rate of 28.2\%, followed by 26.2\% in 2011.
However, a clear improvement in accuracy was achieved by
AlexNet with an error rate of 16.4\%, a 10\%
margin with the runner up. 
AlexNet consists of five convolutional, and three
fully connected layers. The network contains a
total of 61 million parameters. Due to this large size of the network, the
original model had to be trained on two GPUs with a model parallel approach, where the
filters were distributed to these GPUs. Moreover, dropout was required
to avoid overfitting using such a large model size. The next major milestone in
ImageNet classification was made by VGG-Net family~\cite{simonyan2014very},
which exclusively uses $3\times3$ convolutions. The main ideas here were usage of
$3\times3$ convolutions to approximate $7\times7$ filter's receptive
field, along with a deeper network.
However, the model size of VGG-19 with 138 million parameters is even larger
than AlexNet and not suitable for real-time applications.
Another step forward in architecture design was the ResNet family~\cite{he2016deep}, which
incorporates a repetitive structure of $1\times1$ and $3\times3$ convolutions
along with a skip connection. By changing the depth of the networks, the authors
showed competitive performance for multiple learning tasks.

As one can see, a general trend of neural network design has been to find larger and deeper
models to get better accuracy without considering the memory or power
budget. One widely held belief has been that new hardware is going to provide adequate computational power and memory to allow these
networks to run with real-time performance in embedded systems.
However, increase in transistor speed due to semiconductor process improvements has slowed dramatically, and it seems unlikely that mobile processors will meet computational requirements on a limited power budget.
This has opened
several new directions to reduce the memory footprint of existing neural
network architectures using compression~\cite{han2015deep}, 
or  designing new smaller models from scratch. 
SqueezeNet
is a successful example for the latter approach~\cite{iandola2016squeezenet}, which achieves
AlexNet's accuracy with $50\times$ fewer parameters without compression, and $500\times$ smaller with deep compression.
Models for other applications such as detection and segmentation have been developed based on SqueezeNet~\cite{wu2016squeezedet,squeezeseg}.
Another notable work in this direction is the DarkNet Reference network~\cite{darknet}, which
achieves AlexNet's accuracy with $10\times$ fewer parameters (28MB), but
requires $2.8\times$ smaller FLOPs per inference image. They also proposed
a smaller network called TinyDarkNet, which matches AlexNet's performance with only 4.0MB parameters.
Another notable work is MobileNet~\cite{howard2017mobilenets} which used depth wise convolution for spatial convolutions, and was
able to exceed AlexNet's performance with only 1.32 million parameters.
A later work is ShuffleNet~\cite{hluchyj1991shuffle} which extends this idea to pointwise group convolution along with channel shuffling~\cite{hluchyj1991shuffle}. 
More compact versions of Residual Networks have also been proposed. Notable works here include DenseNet~\cite{huang2016densely} and its follow up CondenseNet~\cite{huang2017condensenet}.

\subparagraph{Contributions}
Aiming to design a family of deep neural networks for embedded applications with limited power and memory budgets we present \textit{SqueezeNext}. 
With the smallest version, we can achieve AlexNet's accuracy
with only 0.5 Million model parameters, $112\times$ less than AlexNet
(and more than $2\times$ smaller than SqueezeNet). 
Furthermore, we show how variations
of the network, in terms of width and depth, can span a wide of range of accuracy levels. For instance,
a deeper variation of SqueezeNext can reach VGG-19's baseline accuracy with only 4.4 Million model parameters.
SqueezeNext uses SqueezeNet architecture as a baseline, however we make the following changes.
(i) We use a more aggressive channel reduction by incorporating a two-stage squeeze module. This
significantly reduces the total number of parameters used with the $3\times3$ convolutions. (ii)
We use separable $3\times3$ convolutions to further reduce the model size, and remove the additional $1\times1$ branch after the squeeze module.
(iii) We use an element-wise addition skip connection similar to that of ResNet architecture~\cite{he2016deep}, which allows
us to train much deeper network without the vanishing gradient problem.
In particular, we avoid using DenseNet~\cite{huang2016densely} type connections as it increases the number of channels and
requires concatenating activations which is costly both in terms of execution time and power consumption.
(iv) We optimize the baseline SqueezeNext architecture by simulating its performance on a multi-processor embedded
system. The simulation results gives very interesting insight into where  the bottlenecks for performance are. Based on these observations, one can then perform variations on the baseline model and achieve higher performance both in term of inference speed and power consumption, and sometimes even better classification accuracy.

\begin{figure*}[!htbp]
  \centering
  \includegraphics[width=1.0\textwidth]{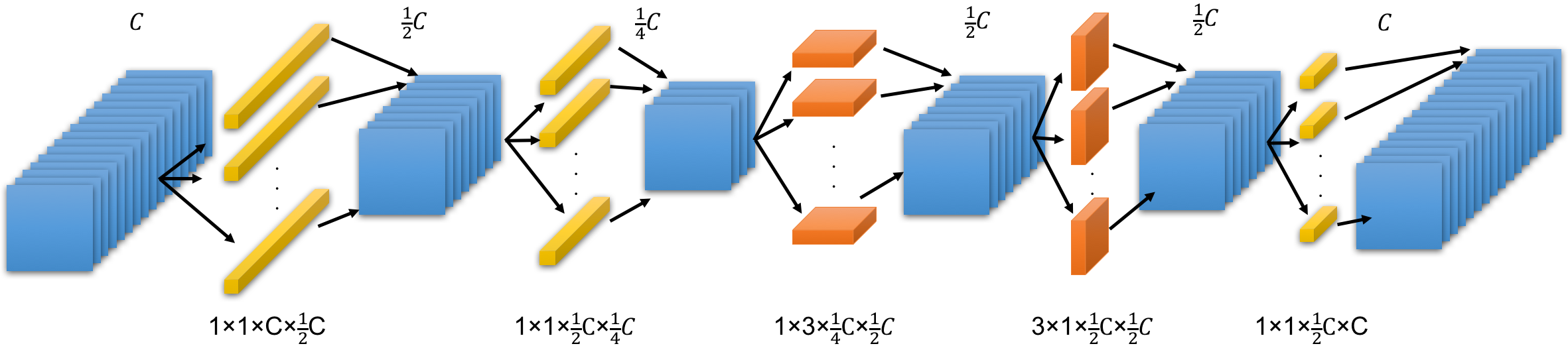}
  \caption{
  Illustration of a SqueezeNext block. An input with $C$ channels is passed through a two stage bottleneck module.
  Each bottleneck module consists of $1\times1$ convolutions reducing the input channel size by a factor of 2. The output
  is then passed through a separable $3\times3$ convolution. The order of $1\times3$ and $3\times1$ convolutions is changed throughout the
  network. The output from the separable convolution is finally passed through an expansion module to match the skip connection's channels
  (the skip connection is not shown here).
  }
  \label{f:sqnxt}
\end{figure*}

\begin{table}[!htbp]
\centering
\caption {
  Performance of the baseline SqueezeNext on ImageNet.
  We report the top-1/top-5 accuracy,
  the number of parameters (\textit{\# Params}), and compression relative to AlexNet (\textit{Comp}).
  The 23 module architecture (1.0-SqNxt-23), exceeds AlexNet's top-5 by $2\%$ margin.
  A more aggressive parameter reduction on this network with 
  group convolutions, is able
  to match AlexNet's top-5 with $112\times$ fewer parameters.
  We also show the performance of deeper variations of the base model with/without Iterative Deep Aggregation (IDA)~\cite{yu2017deep}.
  The 1.0-SqNxt-44 model has the same number of parameters as SqueezeNet but achieves $5\%$ better top-5.
}
\label{t:alexnet}
\begin{tabular}{l|c|c|c|c}
\toprule
Model                     &   Top-1       &   Top-5         &  \# Params        &  Comp.            \\ 
\midrule
  \ga AlexNet             &   57.10       &   80.30         &    60.9M          &  $1\times$            \\ 
  \gc SqueezeNet          &   {57.50}     &   {80.30}       &    {1.2M}         &  $51\times$           \\ 
  \ga 1.0-SqNxt-23        &   59.05       &   82.60         &    0.72M          &  $84\times$           \\ 
  \gc 1.0-G-SqNxt-23      &   \bf{57.16}  &   \bf{80.23}    &    \bf{0.54M}     &  $\bf 112\times$      \\ 
\midrule
  \ga 1.0-SqNxt-23-IDA    &   60.35       &   83.56       &    0.9M             &   $68\times$          \\ 
  \gc 1.0-SqNxt-34        &   61.39       &   84.31       &    1.0M             &   $61\times$          \\ 
  \ga 1.0-SqNxt-34-IDA    &   62.56       &   84.93       &    1.3              &   $47\times$          \\ 
  \gc 1.0-SqNxt-44        &   \bf{62.64}  &   \bf{85.15}  &    \bf{1.2M}        &   $51\times$          \\ 
  \ga 1.0-SqNxt-44-IDA    &   63.75       &   85.97       &    1.5M             &   $41\times$          \\ 
\midrule
\end{tabular}
\end{table}

\section{SqueezeNext Design}
It has been found that many of the filters in the network contain redundant parameters,
in the sense that compressing them would not hurt accuracy. Based on this observation, there has been
many successful attempts to compress a trained network~\cite{jaderberg2014speeding,han2015deep,rastegari2016xnor,zhu2016trained}.
However, some of these compression methods require variable bit-width ALUs for efficient execution.
To avoid this, we aim to design a small network which can be trained from scratch
with few model parameters to start with. To this end, we use the following strategies:

\subparagraph{Low Rank Filters}
We assume that the input to the $i^\text{th}$ layer of the network with $K\times K$ convolution filters to be
$x\in\R^{H\times W\times C_i}$, producing an output activation of $y\in\R^{H\times W\times C_o}$ 
(for ease of notation, we assume that the input and output activations have the same spatial size).
Here, $C_i$ and $C_o$ are the input and output channel sizes. The total number of parameters in this layer
will then be $K^2C_iC_o$. Essentially, the filters would consist of $C_o$ tensors of size $K\times K\times C_i$.

In the case of post-training compression, one seeks to shrink the parameters, $W$, using a low rank basis, $\tilde{W}$.
Possible candidates for $\tilde{W}$ can be CP or Tucker decomposition.
The amount of reduction that can be achieved with these methods is proportional to the rank of the original weight, $W$.
However, examining the trained weights for most of the networks, one finds that they do not have a low rank structure.
Therefore, most works in this area have to perform some form of retraining to recover accuracy~\cite{han2015deep,han2015learning,park2017weighted,hubara2016quantized,rastegari2016xnor}.
This includes pruning to reduce the number of non-zero weights, and reduced precision for elements of $W$.
An alternative to this approach is to re-design a network using the low rank decomposition $\tilde{W}$ to force
the network to learn the low rank structure from the beginning, which is the approach that we follow.
The first change that we make, is to decompose the $K\times K$ convolutions into two separable convolutions of size
$1\times K$ and $K\times 1$, as shown in \fref{f:separable}.
This effectively reduces the number of parameters from $K^2$ to $2K$, and also
increases the depth of the network. These two convolutions
both contain a ReLu activation as well as a batch norm layer~\cite{ioffe2015batch}.

\subparagraph{Bottleneck Module} Other than a low rank structure, the multiplicative factor of $C_i$ and $C_o$ significantly increases the
number of parameters in each convolution layer.
Therefore, reducing the number of input channels would reduce network size.
One idea would be to use depth-wise separable convolution to reduce this multiplicative factor, but this
approach does not good performance on some embedded systems due to its low arithmetic intensity (ratio of compute to bandwidth).
Another ideas is the one used in the SqueezeNet architecture~\cite{iandola2016squeezenet},
where the authors used a squeeze layer before the $3\times3$ convolution to reduce the number of input channels to it.
Here, we use a variation of the latter approach by using a two stage squeeze layer, as shown in~\fref{f:separable}.
In each SqueezeNext block, we use two bottleneck modules each reducing the channel
size by a factor of 2, which is followed by two separable convolutions. We also incorporate a final $1\times1$ expansion module, which
further reduces the number of output channels for the separable convolutions.

\subparagraph{Fully Connected Layers}
In the case of AlexNet, the majority of the network parameters are in Fully Connected layers, accounting for 96\% of the total model size.
Follow-up networks such as ResNet or SqueezeNet consist of only one fully connected layer. Assuming
that the input has a size of $H\times W\times C_i$, then a fully connected
layer for the last layer will contain $H\times W\times C_i\times L$ parameters, where $L$ is the number of labels (1000 for ImageNet).
SqueezeNext incorporates a final bottleneck layer to reduce the input channel size to the last fully
connected layer, which considerably reduces the total number of model parameters. This idea was also used in Tiny DarkNet
to reduce the number parameters~\cite{darknet}.

\begin{figure*}[!htbp]
  \centering
  \includegraphics[width=0.8\textwidth]{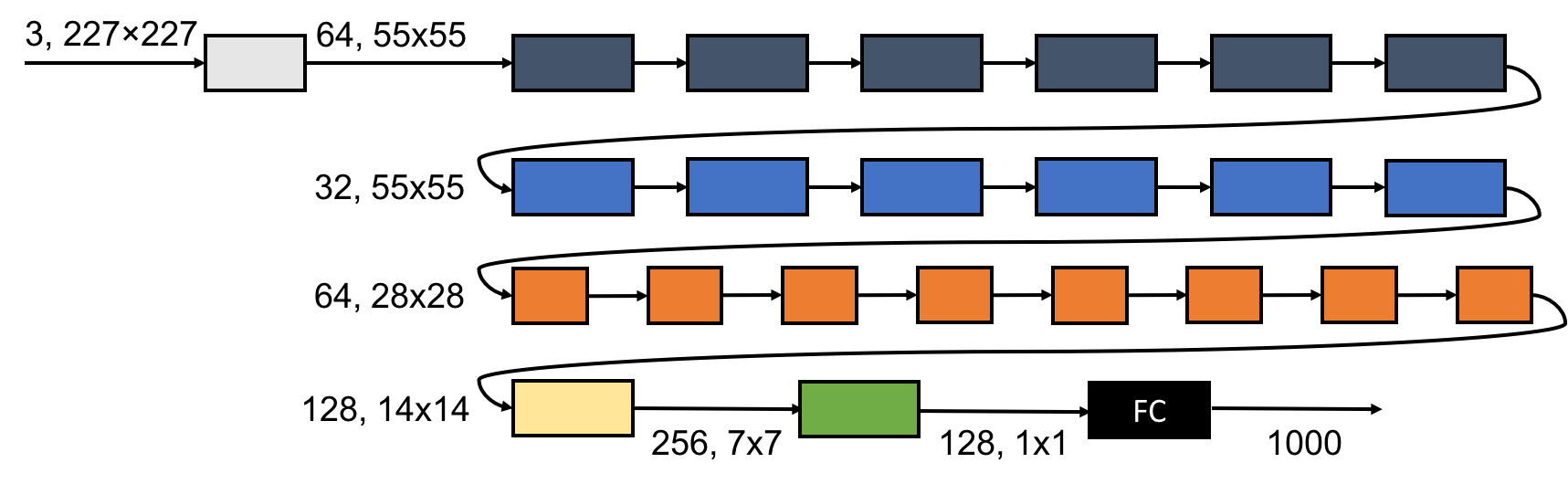}
  \caption{
  Illustration of block arrangement in 1.0-SqNxt-23. Each color change corresponds to a change in input feature
  map's resolution. The number of blocks after the first convolution/pooling layer is $Depth=[6,6,8,1]$, where the last
  number refers to the yellow box. This block is followed by a bottleneck module with average pooling to reduce the channel size and spatial resolution (green box), followed by a fully connected layer (black box). In optimized variations of the baseline, we change this depth distribution by decreasing the number of blocks in early stages (dark blue), and instead assign more blocks to later stages~(\fref{f:v5}). This increases hardware performance as early layers have poor compute efficiency.
  }
  \label{f:sqnxt_depth}
\end{figure*}

\section{Hardware Performance Simulation}
Up to now, hardware architectures have been designed to be optimal for a fixed
neural network, for instance SqueezeNet. 
However, as we later discuss there is important insights that can be gained
from hardware simulation results. This in turn can be used to modify the neural network architecture,
to get better performance in terms of inference and power consumption possibly without incurring generalization loss.
In this section, we first explain how we simulate the performance of the network on a hypothetical neural network accelerator for mobile/embedded systems, and then discuss how the baseline model can be varied to get considerably better hardware performance.

The neural network accelerator is a domain specific processor which is designed to accelerate neural
network inference and/or training tasks. It usually has a large number of computation units
called processing element (PE) and a hierarchical structure of memories and interconnections
to exploit the massive parallelism and data reusability inherent in the convolutional layers.

\SetInd{0.2em}{0.1em}
\SetAlgoNoLine

\begin{algorithm}[!htbp]
\small
\SetKwData{Left}{left}
\SetKwData{This}{this}
\SetKwData{Up}{up}
\SetKwFunction{Union}{Union}
\SetKwFunction{FindCompress}{FindCompress}
\SetKwInOut{Input}{input}
\SetKwInOut{Output}{output}
\Input{Input feature map, $I$, convolution parameters $W$}
\Output{Output feature map, $O$}
\BlankLine
\For(\cmt{Output Channels}){$k\leftarrow 0$ \KwTo $C_o$}{\vspace{-3mm}
  \For(\cmt{H: Height}){$y\leftarrow 0$ \KwTo $H$}{\vspace{-3mm}
    \For(\cmt{W: Width}){$x\leftarrow 0$ \KwTo $W$}{\vspace{-3mm}
      { O[k][y][x]} = 0\;\vspace{1mm}
        \For(\cmt{Input Channels}){$c\leftarrow 0$ \KwTo $C_i$}{\vspace{-3mm}
          \For(\cmt{Filter Size}){$j\leftarrow 0$ \KwTo $K_h$}{\vspace{-3mm}
            \For{$i\leftarrow 0$ \KwTo $K_w$}{
              O[k][y][x] += I[c][y+j][x+i] * W[k][c][j][i]
            }
          }
        }
    }
  }
}
\caption{Execution flow for computing a $K_w\times K_h$
convolution kernel}
\label{a:kiseok}
\end{algorithm}

Eyeriss~\cite{chen2017eyeriss} introduced a taxonomy for classifying neural
network accelerators based on the spatial architecture according to the type of
the data to be reused at the lowest level of the memory hierarchy. This is
related to the order of the six loops of the convolutional layer as shown in
Algorithm~\ref{a:kiseok}.\footnote{Because the typical size of the batch in the
inference task is one, the batch loop is omitted here.}
In our simulator, we consider two options
to execute a convolution: Weight Stationary (WS)
and Output Stationary (OS).

\begin{figure}[!htbp]
  \centering
  \includegraphics[width=0.38\textwidth]{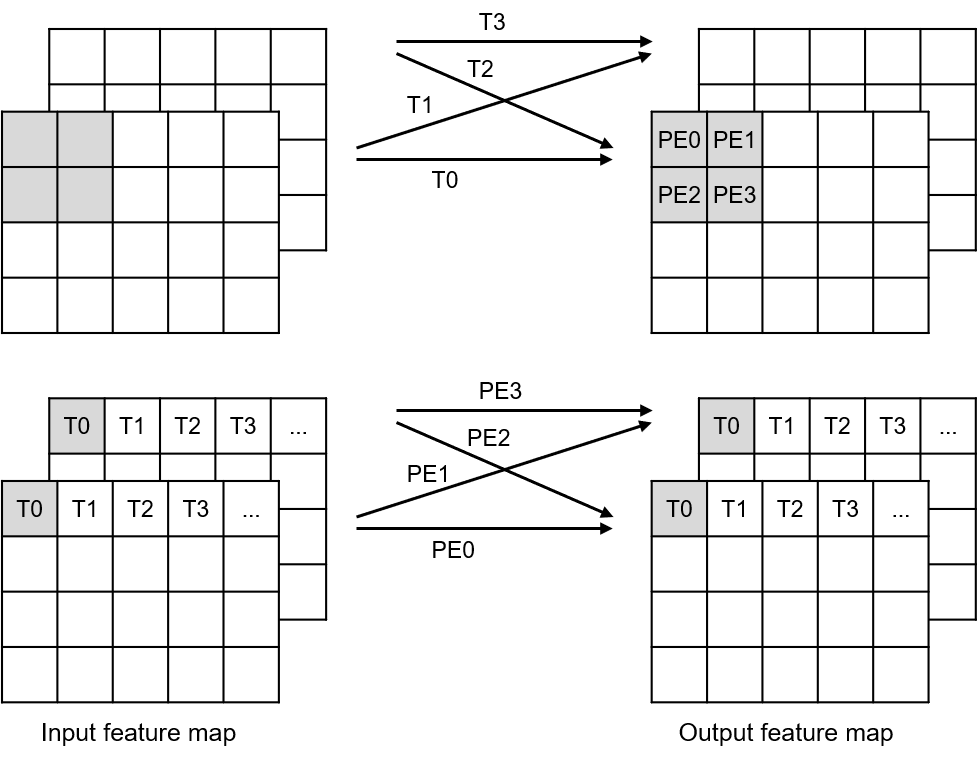}
  \caption{
  Conceptual diagram of two data flows used in the experiment: Output Stationary (top) and Weight Stationary (bottom).
  A $2\times2$ PE array performs a $1\times1$ convolution on a 5x4x2 input (left) generating a 5x4x2 output (right).
  Here $Ti$ denotes the ith cycle. (Top) In T0, and T1 cycles of OS data flow, the shaded area on left is read and convolved with different filter weights and the results are stored in the corresponding output pixels. Then in T2 and T3 cycles the data from the second input channel are read and similar operation is performed to accumulate partial sums. 
  (bottom) in WS the input pixel is first broadcast to the PEs. In the first cycle, PE0 and PE1 apply different convolutions to the first pixel of first input channel and accumulate the results from PE2 and PE3, respectively, and store the results to the corresponding output pixel. In next cycles other input pixels are read and the same operation is performed.
  }
  \label{f:dataflow}
\end{figure}

\subparagraph{Weight Stationary} WS is the most common method used in many notable neural network accelerators~\cite{farabet2009cnp,farabet2011neuflow,gokhale2014240,jouppi2017datacenter,cavigelli2017origami}. For WS method, each PE loads a
convolution filter and executes the convolution at all spatial locations of the input. Once all the spatial positions are iterated, the PE will load the next convolution filter. In this method, the filter weights are
stored in the PE register. For a $H\times W\times C_i$ input feature map,
and $C_o$ $K\times K\times C_i$ convolutions (where $C_o$ is the number of filters),
the execution process is as follows. The PE loads a single
element from the convolution parameters to its local register and applies that to
the whole input activation. Afterwards, it moves to the next element and so
forth. For multiple PEs, we decompose it to a 2D grid of $P_r\times P_c$ processors,
where the $P_r$ dimension decomposes the channels and the $P_c$ dimension
decomposes the output feature maps, $C_o$, as shown in Figure~\ref{f:dataflow}. In summary, in the WS mode the whole
PE array keeps a $P_r\times P_c$ sub-matrix of the weight tensor, and it performs
matrix-vector multiplications on a series of input activation vectors.

\subparagraph{Output Stationary} In the OS method, each PE exclusively works
on one pixel of the output activation map at a time. In each cycle, it applies parts of
the convolution that will contribute to that output pixel, and accumulates the results.
Once all the computations for that pixel are finished, the PE moves to work on
a new pixel. In case of multiple PEs, each processor simply works on different
pixels from multiple channels.  In summary, in the OS mode the whole array
computes a $P_r\times P_c$ block of an output feature map over time. In each cycle, new
inputs and weights needed to compute the corresponding pixel are
provided to each PE. There are multiple ways that OS method could be executed.
These include Single Output Channel-Multiple Output Pixel (SOC-MOP), Multiple
Output Channels-Single Output Pixel (MOC-SOP), and Multiple Output
Channels-Multiple Output Pixels (MOC-MOP)~\cite{chen2017eyeriss}. Here we use
SOC-MOP format.

Accelerators that adopt the weight stationary (WS) data flow are designed to
minimize the memory accesses for the convolution parameters by reusing the weights of the
convolution filter over several activations. On the other hand, accelerators
that use the output stationary (OS) data flow are designed to minimize the
memory access for the output activations by accumulating the partial sums
corresponding to the same output activation over time. Therefore, the $x$ and $y$
loops form the innermost loop in the WS data flow, whereas the $c$, $i$, and $j$ loops
form the innermost loop in the OS data flow. Both data flows show good
performance for convolutional layers with $3\times3$ or larger filters. However,
recent trend on the mobile and embedded neural network architecture is the wide
adoption of lightweight building blocks, e.g. $1\times1$ convolutions, which have limited parallelism and data reusability.

\begin{figure}[!htbp]
  \centering
  \includegraphics[width=0.5\textwidth]{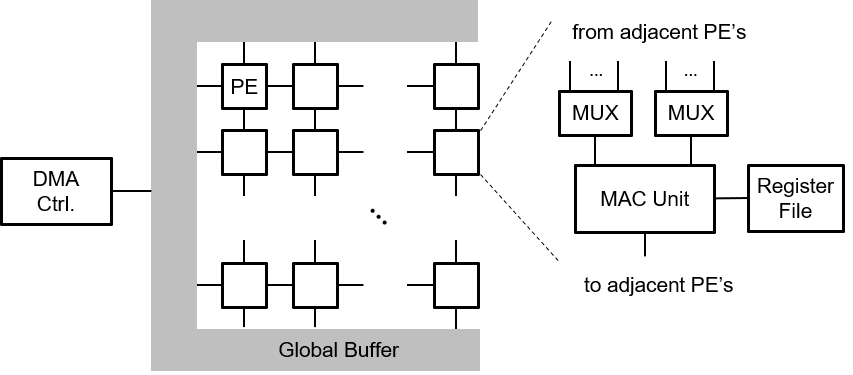}
  \caption{
  Block diagram of the neural network accelerator used as the reference hardware for inference speed and energy estimation of various neural networks. 
  }
  \label{f:accelerator}
\end{figure}

\subparagraph{Hardware Simulation Setup}
Figure~\ref{f:accelerator} shows the block diagram of the neural network
accelerator used as the reference hardware for the inference speed and energy
estimation. It consists of a $16\times16$ or $8\times8$ array of PEs, a 128KB or 32KB
global buffer, and a DMA controller to transfer data between the DRAM and the
buffer. A PE has a 16-bit integer multiply-and-accumulate (MAC) unit and a
local register file. In order for the efficient acceleration of various
configurations of the convolutional layer, the accelerator supports the two
WS and OS operating modes, as explained before. To reduce the execution time and the energy
consumption, the accelerator is designed to exploit the sparsity of the filter
weights~\cite{han2015deep, yang2017designing} in the OS mode as well.
In this
experiment, we conservatively assume 40\% of the weight sparsity.

The accelerator processes the neural network one layer at a time, and the
operating mode which gives better inference speed is used for each layer. The
memory footprint of layers ranges from tens of kilobytes to a few megabytes. If
the memory capacity requirement for a layer is larger than the size of the
global buffer, the tiling is applied to the $x$, $y$, $c$, and $k$ loops of the
convolution in Algorithm~\ref{a:kiseok}. The order and the size of the tiled
loops are selected by using a cost function which takes account of the
inference speed and the number of the DRAM accesses.

The performance estimator computes the number of clock cycles required to
process each layer and sums all the results. The cycles consumed by the PE
array and the global buffer are calculated by modeling the timings of the
internal data paths and the control logic, and the DRAM access time is
approximated by using two numbers, the latency and the effective bandwidth,
which are assumed to be 100 cycles and 16GB/s, respectively. For the energy
estimation, we use a similar methodology to~\cite{chen2017eyeriss}, but the normalized energy cost of components is slightly modified considering the architecture of the reference accelerator.

\begin{table}[!htbp]
\caption {
  Performance of wider variations of the SqueezeNext architecture. The first three rows use $1.5\times$ wider,
  and the last three rows shows $2\times$ wider channels as compared to the respective baseline models.
  The 2.0-SqNxt-44 network is able to match VGG-19's performance with $31\times$ less parameters.
  Furthermore, comparison with MobileNet-1.0-224 shows comparable performance with $1.3\times$ fewer parameters 
  (2.0-SqNxt23v5).
}
\label{t:vgg}
\centering
\begin{tabular}{c|c|c|c}
\midrule
  Model                   &   Top-1   &   Top-5   &  Params \\ 
\midrule
  \ga 1.5-SqNxt-23       &   63.52         &   85.66         &    1.4M         \\ 
  \gc 1.5-SqNxt-34       &   66.00         &   87.40         &    2.1M         \\ 
  \ga 1.5-SqNxt-44       &   67.28         &   88.15         &    2.6M         \\ 
\midrule
  \ga VGG-19             &   \bf{68.50}    &   \bf{88.50}    &   \bf{138M}     \\ 
  \gc 2.0-SqNxt-23       &   67.18         &   88.17         &    2.4M         \\ 
  \ga 2.0-SqNxt-34       &   68.46         &   88.78         &    3.8M         \\ 
  \gc 2.0-SqNxt-44       &   \bf{69.59}    &   \bf{89.53}    &    \bf{4.4M}    \\ 
\midrule
  \ga MobileNet          &   67.50 (70.9)  &   86.59 (89.9)  &   4.2M    \\
\gc 2.0-SqNxt-23v5     &   67.44 (69.8)  & \bf{88.20}(89.5)&   \bf{3.2M}    \\      
\bottomrule
\end{tabular}
\end{table}

\begin{figure*}[!htbp]
  \centering
  \includegraphics[width=0.99\textwidth]{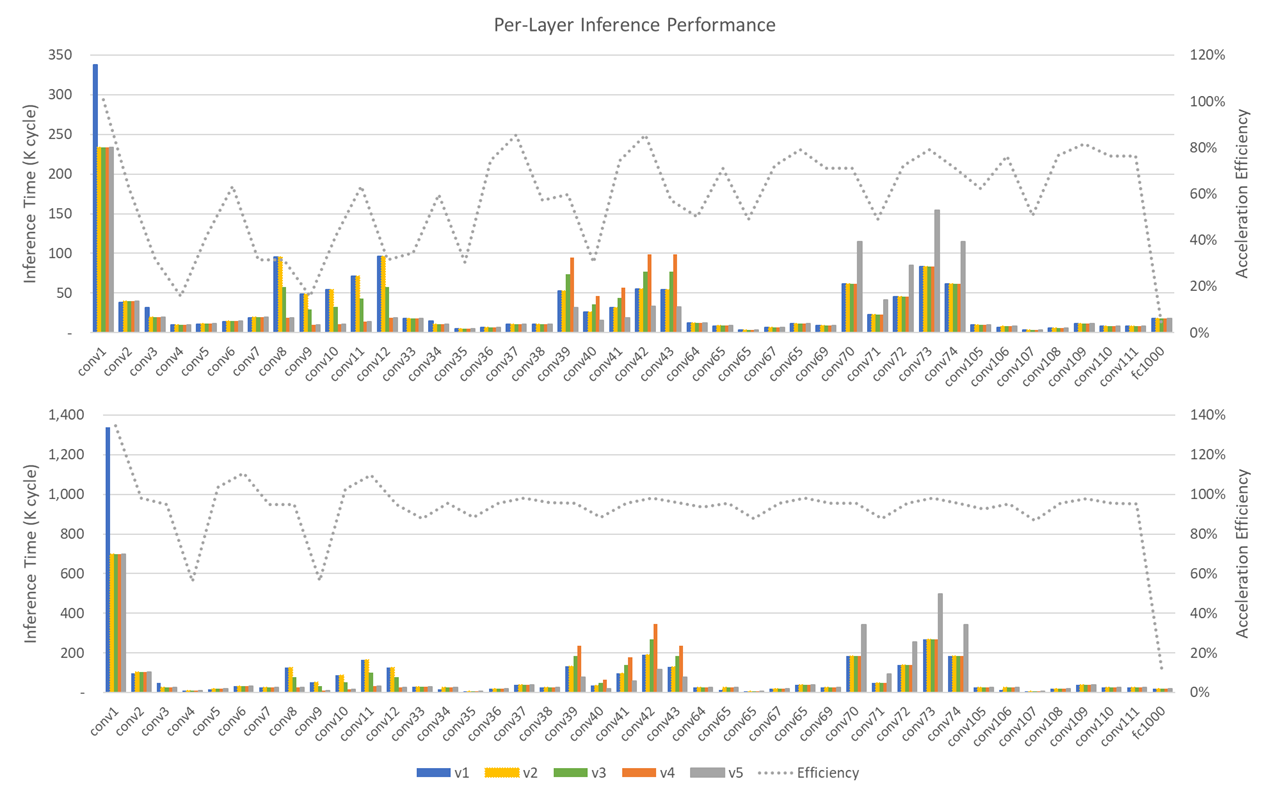}
  \caption{
  Per-layer inference time (lower is better)  is shown  along the left y-axis for
  variants (v1-v5) of 1.0-SqNxt-23 architecture. Acceleration efficiency (number of MAC operations divided by total cycle counts) is shown by the dotted line and the right y-axis. The top graph shows the results for an $16\times16$ array and the bottom a $8\times8$ PE array.
  Note the relatively poor efficiency in early layers for the $16\times16$ PE array due to the small number of
  filter channels.}
  \label{f:cycle_counts}
\end{figure*}

\section{Results}
\label{s:results}

 Embedded applications have a variety of constraints with regard to accuracy, power, energy, and speed. To meet these constraints we explore a variety of trade-offs. Results of these explorations are reported in this section.

\begin{table*}[!htbp]
\caption {
Simulated hardware performance results in terms of inference time and energy for the
$8\times8$ and $16\times16$ PE array configurations. The time for each configuration is normalized by the number of cycles of the fastest
network for each configuration (smaller is better).
Note how the variations of the baseline SqueezeNext model are able to achieve better inference and
power consumption. For instance, the 1.0-SqNxt-23v5 model is 12\% faster and 17\% more energy efficient than the baseline model for $16\times16$ configuration.
This is achieved by an efficient redistribution of depth at each stage (see~\fref{f:sqnxt_depth}).
Also note that the 2.0-SqNxt23-v5 has better energy efficiency as compared to MobileNet.
}
\label{t:estimation2}
\centering
\begin{tabular}{l|r|r|r|r|c|r|r|r|r}
\toprule
  Model               & Params  & MAC &   Top-1 & Top-5       &Depth        & \multicolumn{2}{c|}{8x8, 32KB} & \multicolumn{2}{c}{16x16, 128KB} \\
                      & (\small$\times$~1\pexp{6})&             &             &             &              & Time        & Energy       &  Time     & Energy \\
\midrule
  \ga AlexNet         & 60.9    & 725M & 57.10    &   80.30     & ---         & x5.46       & 1.6E+10      & x8.26       & 1.5E+10 \\
  \ga SqueezeNet v1.0 & 1.2     & 837M & 57.50    &   80.30     & ---         & x3.42       & 6.7E+09      & x2.59       & 4.5E+09 \\
  \gc SqueezeNet v1.1 & 1.2     & 352M & 57.10    &   80.30     & ---         & x1.60       & 3.3E+09      & x1.31       & 2.4E+09 \\
  \ga Tiny Darknet    & 1.0     & 495M & 58.70    &   81.70     & ---         & x1.92       & 3.8E+09      & x1.50       & 2.5E+09 \\
  \ga 1.0-SqNxt-23    & 0.72    & 282M & 59.05    &   82.60     & [6,6,8,1]   & x1.17       & 3.2E+09      & x1.22       & 2.5E+09 \\      %
  \gc 1.0-SqNxt-23v2  & 0.74    & 228M & 58.55    &   82.09     & [6,6,8,1]   & \bf{x1.00}  & 2.8E+09      & x1.13       & 2.4E+09 \\      
  \ga 1.0-SqNxt-23v3  & 0.74    & 228M & 58.18    &   81.96     & [4,8,8,1]   & x1.00       & 2.7E+09      & x1.08       & 2.3E+09 \\      
  \gc 1.0-SqNxt-23v4  & 0.77    & 228M & 59.09    &   82.41     & [2,10,8,1]  & x1.00       & 2.6E+09      & x1.02       & 2.2E+09 \\      
  \ga 1.0-SqNxt-23v5  & 0.94&\bf{228M} &\bf{59.24}&   82.41     & [2,4,14,1]  & x1.00       & \bf{2.6E+09} & \bf{x1.00}  & \bf{2.0E+09} \\ 
\midrule
  \gc MobileNet       & 4.2     & 574M & 67.50(70.9)& 86.59(89.9)& ---         & \bf{x2.94}  & 9.1E+09      & x2.60       & 5.8E+09 \\
  \ga 2.0-SqNxt-23    & 2.4     & 749M & 67.18    &   88.17     & [6,6,8,1]   & x3.24       & 8.1E+09      & x2.72       & 5.9E+09 \\      %
  \gc 2.0-SqNxt-23v4  & 2.56    & 708M & 66.95    &   87.89     & [2,10,8,1]  & x3.17       & 7.5E+09      & x2.55       & 5.4E+09 \\      
  \ga 2.0-SqNxt-23v5  &\bf{3.23}& 708M & 67.44    &   \bf{88.20}& [2,4,14,1]  &    {x3.17}  & \bf{7.4E+09} & \bf{x2.55}       & \bf{5.4E+09} \\      
\bottomrule
\end{tabular}
\end{table*}

\subparagraph{Training Procedure}
For training, we use a  $227\times 227$ center crop of the input image and
subtract the ImageNet mean from each input channel. We do not use any further
data augmentation and use the same hyper-parameters for all the experiments. Tuning hyper-parameter can
increase the performance of some of the SqueezeNext variants, but it is not expected to
change the general trend of our results.
To accelerate training, we use
a data parallel method, where the input batch size is distributed among $P$
processors (hereforth referred to as PE). Unless otherwise noted, we use $P=32$ Intel KNightsLanding (KNL), each
consisting of 68 1.4GHz Cores with a single precision peak of 6TFLOPS.  All
experiments were performed on VLAB system which consists of 256 KNLs, with an
Intel Omni-Path 100 series interconnect.  
We perform a 120 epoch training
using Intel Caffe with a batch size of $B=1024$.
In the data parallel approach, the
model is replicated to all workers and each KNL gets an input batch size of
$B/P$, randomly selected from the training data and independently performs a
forward and backwards pass~\cite{gholami2017integrated}. This is followed by a collective all-reduce
operation, where the sum of the gradients computed through backward pass in each
worker is computed. This information is then broadcasted to each worker and the
model is updated.

\subparagraph{Classification Performance Results}
We report the performance of the SqueezeNext architecture in~\tref{t:alexnet}.
The network name is appended by the number of modules. The schematic of each module is shown in~\fref{f:separable}.
Simply because AlexNet has been widely used as a reference architecture in the literature, we begin with a comparison to AlexNet.
Our 23 module architecture
exceeds AlexNet's performance with a $2\%$ margin with $87\times$ smaller number of parameters. 
Note that in the
SqueezeNext architecture, the majority of the parameters are in the $1\times 1$ convolutions.
To explore how much further we can reduce the size of the network,
we use group convolution with a group size of two.
Using this approach we are able to match AlexNet's top-5 performance with a $112\times$ smaller model.

Deeper variations of SqueezeNext can
actually cover a wide range of accuracies as reported in~\tref{t:alexnet}. The deepest model we tested consists of 44 modules: 1.0-SqNxt-44. 
This model achieves $5\%$ better top-5 accuracy as compared to AlexNet. SqueezeNext modules can also be used
as building blocks for other types of network designs. One such possibility is to use a tree structure instead of the typical feed forward architecture as proposed in~\cite{yu2017deep}. Using the Iterative Deep Aggregation (IDA) structure, we can achieve better accuracy, although it increases the model size.
Another variation for getting better performance is to increase the network width.
We increase the baseline width by a multiplier factor of $1.5$ and $2$ and
report the results in~\tref{t:vgg}. The version with twice the width and 44 modules (2.0-SqNxt-44) is able to match VGG-19's performance with $31\times$ smaller number of parameters.

A novel family of neural networks particularly designed for mobile/embedded applications is MobileNet, which uses depth-wise convolution. Depth-wise convolutions reduce the parameter count, but also have poor arithmetic intensity. 
MobileNet's best reported accuracy results have benefited from data augmentation and extensive experimentation with training regimen and hyper-parameter tuning
(these results are reported in parentheses in~\tref{t:vgg}).
Aiming to perform a fairer comparison with these results with SqueezeNext, we
trained MobileNet under similar training regimen to SqueezeNext and report the results in~\tref{t:vgg}.
SqueezeNext is able to achieve similar results for Top-1 and slightly better Top-5 with half the model parameters.
It may be possible to get better results for SqueezeNext with hyper-parameter tuning which is network specific.
However, our main goal is to show the general performance trend of SqueezeNext and not the maximum achievable performance for each individual version of it.

\subparagraph{Hardware Performance Results}

Figure~\ref{f:cycle_counts} shows the per-layer cycle count estimation of
1.0-SqNxt-23, along with its optimized variations explained below. For better
visibility, the results of the layers with the same configuration are
summed together, e.g. Conv8, Conv13, Conv18, and Conv23, and represented as a
single item. In the 1.0-SqNxt-23, the first $7\times7$ convolutional layer accounts for 26\% of the total
inference time.
This is due to its relatively
large filter size applied on a large input feature map,
Therefore, the first optimization we make is replacing this $7\times7$
layer with a $5\times5$ convolution, and construct 1.0-SqNxt-23-v2 model.
Moreover, we plot the accelerator efficiency in terms of flops per
cycle for each layer of the 1.0-SqNxt-23 model. Note the significant drop in efficiency for
the layers in the first module. This drop is more significant for the $16\times16$ PE array
configuration as compared to the $8\times8$ one. The reason for this drop is that the initial layers have very small
number of channels which needs to be applied a large input activation map. However, later layers
in the network do not suffer from this, as they contain filters with larger channel size. To resolve
this issue, we explore changing the number of modules to better distribute the workload. The baseline model
has $6-6-8-1$ modules before the activation map size is reduced at each stage, as shown in~\fref{f:sqnxt_depth}.
We consider
three possible variations on top of the v2 model. In the v3/v4 variation,
we reduce the number of the blocks in the first module by 2/4 and instead add it to the second module, respectively. In the v5 variation,
we reduce the blocks of the first two modules and instead increase the blocks in the third module.
The results are shown in~\tref{t:estimation2}.
As one can see, the v5 variation (\textit{1.0-SqNxt-23v5})
actually achieves better top-1/5 accuracy and has much better performance on the $16\times16$ PE array. It uses
17\% lower energy and is 12\% faster as compared to the baseline model (i.e. 1.0-SqNxt-23). In total, the latter network is $2.59\times$/$8.26\times$
faster and $2.25\times$/$7.5\times$ more energy efficient as compared to SqueezeNet/AlexNet without any accuracy degradation.

\begin{figure}[!htbp]
  \centering
  \includegraphics[width=0.43\textwidth]{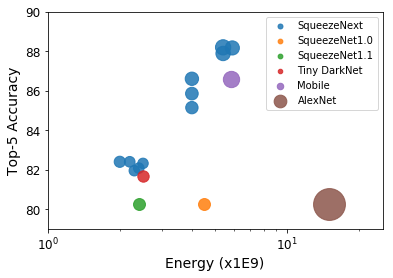}\\
  \includegraphics[width=0.43\textwidth]{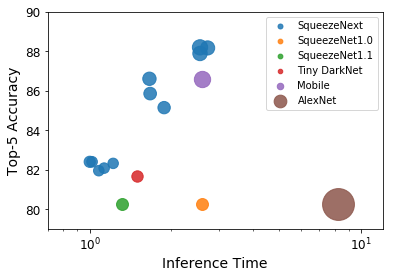}
  \caption{
  The spectrum of accuracy versus energy and inference speed for SqueezeNext, SqueezeNet (v1.0 and v1.1), Tiny DarkNet, and MobileNet. 
  SqueezeNext shows superior performance (in both plots higher and to the left is better).
  The circle areas are proportional to square root of model size for each network.
  }
  \label{f:trend}
\end{figure}

Very similar performance improvement
is observed with the v4 variation, with only 50K higher number of parameters.
We also show results comparing MobileNet and 2.0-SqNxt-23v5 which matches its classification performance.
SqueezeNext has lower energy consumption, and achieves better speedup when we use a $16\times16$ PE array as compared to $8\times8$. The reason for this is the inefficiency of depthwise-separable convolution
in terms of hardware performance, which is due to its poor arithmetic intensity (ratio of compute to memory operations)~\cite{williams2009roofline}.
This inefficiency becomes more pronounced as higher number of processors are used, since the problem
becomes more bandwidth bound. A comparative plot for trade-offs between energy, inference speed, and accuracy for
different networks is shown in~\fref{f:trend}. As one can see, SqueezeNext provides a family of networks
that provide superior accuracy with good power and inference speed.

\section{Conclusions}
\label{s:conclusions}

In this work, we presented \textit{SqueezeNext}, a new family of neural network architectures that is 
able to achieve AlexNet's top-5 performance with $112\times$ fewer parameters.
A deeper variation of the SqueezeNext architecture exceeds VGG-19's accuracy with $31\times$ fewer parameters.
MobileNet is a very novel network for Mobile applications, but SqueezeNext was able to exceed MobileNet's top-5 accuracy by $1.6\%$, with $1.3\times$ fewer parameters.
SqueezeNext accomplished this without using depthwise-separable convolutions that are troublesome for some mobile processor-architectures. 
The baseline network consists of a two-stage bottleneck module to reduce the number of input channels to spatial convolutions,
use of low rank separable convolutions, along with an expansion module. We also restrict the number of input channels to the fully connected
layer to further reduce the model parameters.
More efficient variations of the baseline architecture are achieved by simulating
the hardware performance of the model on the PE array of a realistic neural network accelerator and reported
the results in terms of power and inference cycle counts. 
Using per layer simulation analysis, we proposed
 different variations of the baseline model that can not only achieve better inference speed and power energy consumption,
but also got better classification accuracy with only negligible increase in model size.
The tight coupling between neural net design and performance modeling on a neural net accelerator architecture was essential to get our result.
This allowed us to design a novel network that is $2.59\times$/$8.26\times$ faster and $2.25\times$/$7.5\times$ more energy efficient as compared to SqueezeNet/AlexNet without any accuracy degradation.

The resulting wide of range of speed, energy, model-size, and accuracy trade-offs provided by the SqueezeNext family allows the user to select the right neural net model for a particular application.

\vspace{-5mm}
\subparagraph{Acknowledgements} We would like to thank Forrest Iandola for his
constructive feedback as well as the generous support from Intel (S. Buck, J, Ota, B. Forgeron, and S. Goswami).

\bibliographystyle{ieee}
\bibliography{ref}

\clearpage
\begin{figure*}
  \centering
  \includegraphics[width=0.9\textwidth]{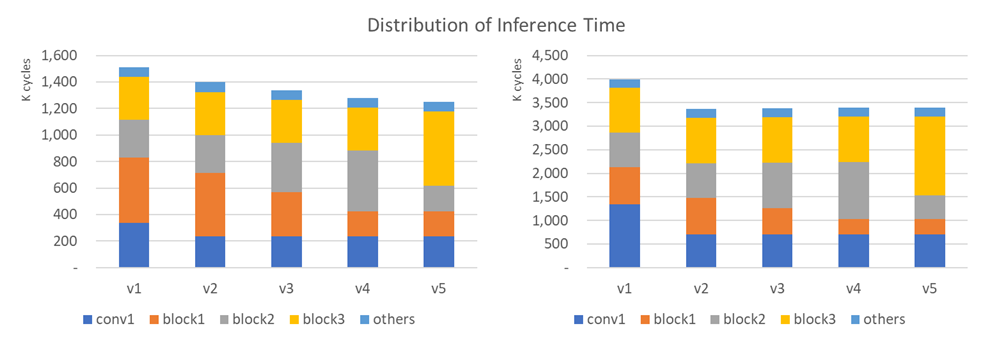}
  \caption{
  Distribution of the inference time for the baseline and its optimized variants is shown, along with the contributions 
  corresponding to different modules. Here the modules are broken into blocks that get the same input resolution (height and width).
  For instance, all layers in block1 have an input feature map of $55\times55$, and block2 gets $28\times28$, etc.
  The results for the  $16\times16$ and $8\times8$ PE configurations are shown in the left and right plots, respectively. 
  }
  \label{f:blk_time}
\end{figure*}

\begin{figure*}
  \centering
  \includegraphics[width=0.9\textwidth]{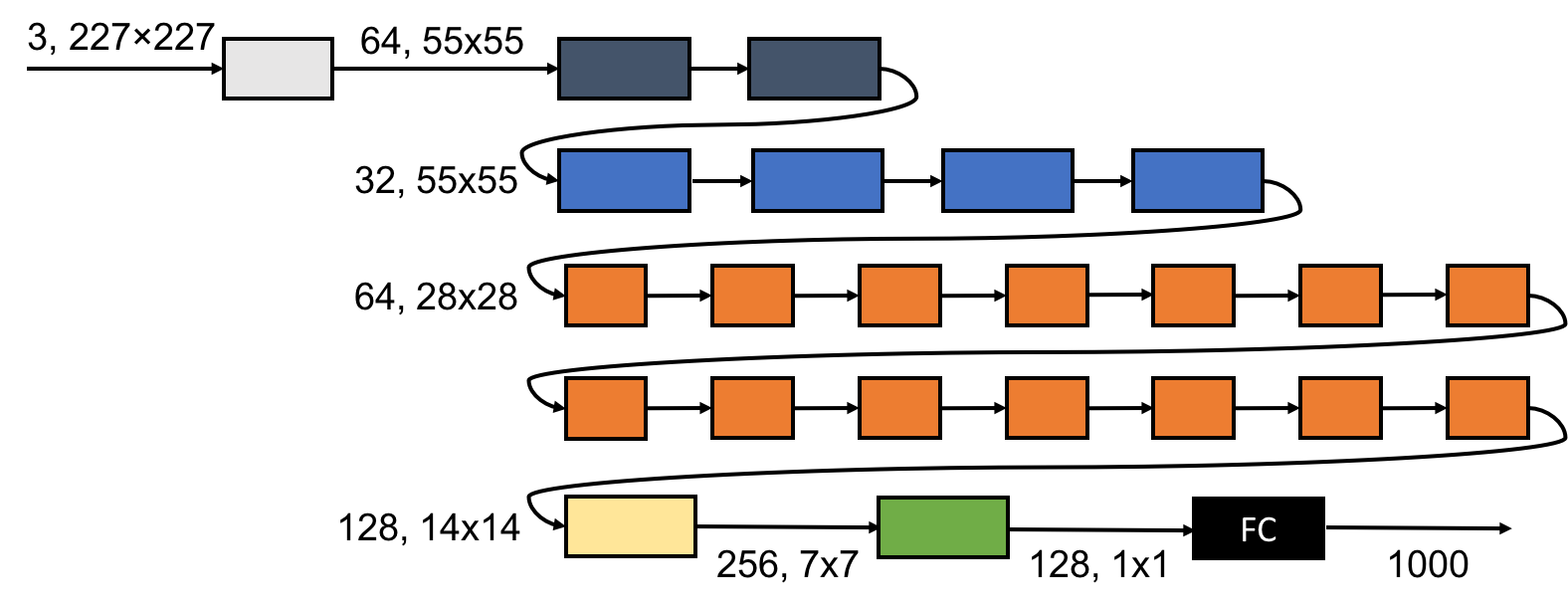}
  \caption{
  Illustration of block arrangement in 1.0-SqNxt-23v5, which is a redistribution of the number of
  modules at each stage to achieve better hardware performance (energy and inference speed).
  Each color change corresponds to a change in input feature
  map's resolution. The number of blocks after the first convolution/pooling layer is $Depth=[2,4,14,1]$, where the last
  number refers to the yellow box. This has to be compared with~\fref{f:sqnxt_depth}, which illustrates
  the baseline configuration.
  }
  \label{f:v5}
\end{figure*}

\begin{table*}
\caption {
Breakdown of the 1.0-SqNxt-23 architecture. Here, $W_i$, $H_i$, and $C_i$ are the input width, height, and channels, $K_w$ and $K_h$ are the
convolution filter sizes,  $W_o$, $H_o$, and $C_o$ are the output width, height, and channels, $S$ is the stride, $P_w$ and $P_h$ is padding,
Repeat refers to how many times a particular module is repeated, Params is the number of parameters for each layer, MAC is the number of multiply and add
instructions, and Act. Size is the size of the output activation in terms of MB. Note that the values for Params and MAC columns are divided by 1000.
}
\label{t:net_config}
\centering
\small
\begin{tabular}{cccccccccccccccccc}
\toprule 
\ga Name		&Type	&$W_i$ 	&$H_i$ 	&$C_i$ 	&$K_w$ 	&$K_h$ 	&$W_o$ 	&$H_o$ 	&$C_o$ 	&$S$	&$P_w$ 	&$P_h$ 	&Repeat	 	&Params 	&	MAC 		&Act. Size\\ 
\midrule
\gc conv1		&Conv	&227 	&227 	&3		&7		&7		&111 	&111	&64		&2		&0		&0		&1		&9.5 	& 116,705  	&3080	\\		
\ga conv2		&Conv	&55		&55		&64		&1		&1		&55		&55		&32		&1		&0		&0		&1		&2.1 	& 6,292  	&378	\\		
\gc conv3		&Conv	&55		&55		&64		&1		&1		&55		&55		&16		&1		&0		&0		&1		&1.0 	& 3,146  	&189	\\		
\ga conv4		&Conv	&55		&55		&16		&1		&1		&55		&55		&8		&1		&0		&0		&1		&0.1 	& 411  		&95		\\	
\gc conv5		&Conv	&55		&55		&8		&1		&3		&55		&55		&16		&1		&0		&1		&1		&0.4 	& 1,210  	&189	\\		
\ga conv6		&Conv	&55		&55		&16		&3		&1		&55		&55		&16		&1		&1		&0		&1		&0.8 	& 2,372  	&189	\\			
\gc conv7		&Conv	&55		&55		&16		&1		&1		&55		&55		&32		&1		&0		&0		&1		&0.5 	& 1,646  	&378	\\				
\midrule
\ga conv8		&Conv	&55		&55		&32		&1		&1		&55		&55		&16		&1		&0		&0		&5		&2.6 	& 7,986  	&189	\\		
\gc conv9		&Conv	&55		&55		&16		&1		&1		&55		&55		&8		&1		&0		&0		&5		&0.7 	& 2,057  	&95		\\	
\ga conv10		&Conv	&55		&55		&8		&3		&1		&55		&55		&16		&1		&1		&0		&5		&2.0 	& 6,050  	&189	\\		
\gc conv11		&Conv	&55		&55		&16		&1		&3		&55		&55		&16		&1		&0		&1		&5		&3.9 	& 11,858  	&89		\\
\ga conv12		&Conv	&55		&55		&16		&1		&1		&55		&55		&32		&1		&0		&0		&5		&2.7 	& 8,228  	&378	\\	
\midrule
\gc conv33		&Conv	&55		&55		&32		&1		&1		&28		&28		&64		&2		&0		&0		&1		&2.1 	& 1,656  	&196	\\	
\ga conv34		&Conv	&55		&55		&32		&1		&1		&28		&28		&32		&2		&0		&0		&1		&1.1 	& 828  		&98		\\
\gc conv35		&Conv	&28		&28		&32		&1		&1		&28		&28		&16		&1		&0		&0		&1		&0.5 	& 414  		&49		\\
\ga conv36		&Conv	&28		&28		&16		&1		&3		&28		&28		&32		&1		&0		&1		&1		&1.6 	& 1,229  	&98		\\
\gc conv37		&Conv	&28		&28		&32		&3		&1		&28		&28		&32		&1		&1		&0		&1		&3.1 	& 2,434  	&98		\\
\ga conv38		&Conv	&28		&28		&32		&1		&1		&28		&28		&64		&1		&0		&0		&1		&2.1 	& 1,656  	&196	\\	
\midrule
\gc conv39		&Conv	&28		&28		&64		&1		&1		&28		&28		&32		&1		&0		&0		&5		&10.4 	& 8,154  	&98		\\
\ga conv40		&Conv	&28		&28		&32		&1		&1		&28		&28		&16		&1		&0		&0		&5		&2.6 	& 2,070  	&49		\\
\gc conv41		&Conv	&28		&28		&16		&3		&1		&28		&28		&32		&1		&1		&0		&5		&7.8 	& 6,147  	&98		\\
\ga conv42		&Conv	&28		&28		&32		&1		&3		&28		&28		&32		&1		&0		&1		&5		&15.5 	& 12,168  	&98		\\	
\gc conv43		&Conv	&28		&28		&32		&1		&1		&28		&28		&64		&1		&0		&0		&5		&10.6 	& 8,279  	&196	\\	
\midrule
\ga conv64		&Conv	&28		&28		&64		&1		&1		&14		&14		&128	&2		&0		&0		&1		&8.3 	& 1,631  	&98		\\
\gc conv65		&Conv	&28		&28		&64		&1		&1		&14		&14		&64		&2		&0		&0		&1		&4.2 	& 815  		&49		\\
\ga conv66		&Conv	&14		&14		&64		&1		&1		&14		&14		&32		&1		&0		&0		&1		&2.1 	& 408  		&25		\\
\gc conv67		&Conv	&14		&14		&32		&1		&3		&14		&14		&64		&1		&0		&1		&1		&6.2 	& 1,217  	&49		\\
\ga conv68		&Conv	&14		&14		&64		&3		&1		&14		&14		&64		&1		&1		&0		&1		&12.4 	& 2,421  	&49		\\
\gc conv69		&Conv	&14		&14		&64		&1		&1		&14		&14		&128	&1		&0		&0		&1		&8.3 	& 1,631  	&98		\\	
\midrule
\ga conv70		&Conv	&14		&14		&128	&1		&1		&14		&14		&64		&1		&0		&0		&7		&57.8 	& 11,327  	&49		\\
\gc conv71		&Conv	&14		&14		&64		&1		&1		&14		&14		&32		&1		&0		&0		&7		&14.6 	& 2,854  	&25		\\
\ga conv72		&Conv	&14		&14		&32		&3		&1		&14		&14		&64		&1		&1		&0		&7		&43.5 	& 8,517  	&49		\\
\gc conv73		&Conv	&14		&14		&64		&1		&3		&14		&14		&64		&1		&0		&1		&7		&86.5 	& 16,947  	&49		\\	
\ga conv74		&Conv	&14		&14		&64		&1		&1		&14		&14		&128	&1		&0		&0		&7		&58.2 	& 11,415  	&98		\\
\midrule
\gc conv105		&Conv	&14		&14		&128	&1		&1		&7		&7		&256	&2		&0		&0		&1		&33.0 	& 1,618  	&49		\\
\ga conv106		&Conv	&14		&14		&128	&1		&1		&7		&7		&128	&2		&0		&0		&1		&16.5 	& 809  		&25		\\
\gc conv107		&Conv	&7		&7		&128	&1		&1		&7		&7		&64		&1		&0		&0		&1		&8.3 	& 405  		&12		\\
\ga conv108		&Conv	&7		&7		&64		&1		&3		&7		&7		&128	&1		&0		&1		&1		&24.7 	& 1,210  	&25		\\
\gc conv109		&Conv	&7		&7		&128	&3		&1		&7		&7		&128	&1		&1		&0		&1		&49.3 	& 2,415  	&25		\\
\ga conv110		&Conv	&7		&7		&128	&1		&1		&7		&7		&256	&1		&0		&0		&1		&33.0 	& 1,618  	&49		\\
\gc conv111		&Conv	&7		&7		&256	&1		&1		&7		&7		&128	&1		&0		&0		&1		&32.9 	& 1,612  	&25		\\
\ga fc1000		&FC		&1		&1		&128	&1		&1		&1		&1		&1000	&1		&0		&0		&1		&129.0 	& 129  		&4		\\
\bottomrule
\end{tabular}
\end{table*}

\end{document}